\documentclass[journal,twoside,web]{ieeecolor}
\usepackage{tmi}
\usepackage{cite}
\usepackage{amsmath,amssymb,amsfonts}
\usepackage{algorithmic}
\usepackage{subfigure}
\usepackage{graphicx}
\usepackage{textcomp}
\usepackage{hyperref}
\hypersetup{
    colorlinks=true,
    linkcolor=blue,  
    citecolor=blue,  
    urlcolor=blue,   
    linkbordercolor=0 0 0, 
}
\usepackage{siunitx}  
\usepackage{booktabs} 
\usepackage{multirow}    
\usepackage{booktabs}    
\usepackage{siunitx}     
\usepackage[table]{xcolor} 

\def\BibTeX{{\rm B\kern-.05em{\sc i\kern-.025em b}\kern-.08em
    T\kern-.1667em\lower.7ex\hbox{E}\kern-.125emX}}
\begin{document}
\title{CEPerFed: Communication-Efficient Personalized Federated Learning for Multi-Pulse MRI Classification}
\author{{Ludi Li, Junbin Mao, Hanhe Lin, Xu Tian, Fang-Xiang Wu, Jin Liu}
\thanks{This work was supported in part by the National Natural Science Foundation of China under Grant 62172444 and Grant 62472450, in part by the Central South University Innovation-Driven Research Programme under Grant 2023CXQD018, and in part by the High Performance Computing Center of Central South University. (Ludi Li and Junbin Mao are equally contributed to this work. Corresponding author: Jin Liu.)}
\thanks{Ludi Li is with Xinjiang Engineering Research Center of Big Data and Intelligent Software, School of Software, Xinjiang University, Urumqi 830091, China (e-mail:liludi@stu.xju.edu.cn).}
\thanks{Junbin Mao and Xu Tian are with Hunan Province Key Laboratory on Bioinformatics, School of Computer Science and Engineering, Central South University, Changsha 410083, China (e-mail:maojunbin@csu.edu.cn; tianxu@csu.edu.cn).}
\thanks{Hanhe Lin is with School of Science and Engineering, University of Dundee, DD1 4HN Dundee, United Kingdom (e-mail: hlin001@dundee.ac.uk).}
\thanks{Fang-Xiang Wu is with Division of Biomedical Engineering, University of Saskatchewan, Saskatoon SK S7N 5A9, Canada (e-mail: fangxiang.wu@usask.ca).}
\thanks{Jin Liu is with Hunan Provincial Key Laboratory on Bioinformatics, School of Computer Science and Engineering, Central South University, Changsha 410083, China, and also with Xinjiang Engineering Research Center of Big Data and Intelligent Software, School of Software, Xinjiang University, Urumqi 830091, China (e-mail: liujin06@csu.edu.cn).}}

\maketitle

\begin{abstract}
Multi-pulse magnetic resonance imaging (MRI) is widely utilized for clinical practice such as Alzheimer's disease diagnosis. To train a robust model for multi-pulse MRI classification, it requires large and diverse data from various medical institutions while protecting privacy by preventing raw data sharing across institutions. Although federated learning (FL) is a feasible solution to address this issue, it poses challenges of model convergence due to the effect of data heterogeneity and substantial communication overhead due to large numbers of parameters transmitted within the model. To address these challenges, we propose CEPerFed, a communication-efficient personalized FL method. It mitigates the effect of data heterogeneity by incorporating client-side historical risk gradients and historical mean gradients to coordinate local and global optimization. The former is used to weight the contributions from other clients, enhancing the reliability of local updates, while the latter enforces consistency between local updates and the global optimization direction to ensure stable convergence across heterogeneous data distributions. To address the high communication overhead, we propose a hierarchical SVD (HSVD) strategy that transmits only the most critical information required for model updates. Experiments on five classification tasks demonstrate the effectiveness of the CEPerFed method.
The code will be released upon acceptance at https://github.com/LD0416/CEPerFed.


\end{abstract}


\section{Introduction}
\label{sec:introduction}

\IEEEPARstart{M}{agnetic} resonance imaging (MRI) is a noninvasive, radiation-free technology that is widely used for disease diagnosis and clinical monitoring. A MRI pulse is a systematically coordinated series of medical signals such as radiofrequency pulses, gradient pulses, signal acquisition timings, or their combinations, engineered for differentiating anatomical structures by generating tissue-specific contrast \cite{calle2018basic, sharma2019missing}. Models trained on multi-pulse MRI can effectively utilize their complementary information and improve the accuracy of classification~\cite{kaufmann2020consensus}. Training a such model with good generliability requires large-scale datasets, however, data in a single institution is often limited. Alternatively, training a model on data collected from multiple medical institutions could solve this problem but violate patient privacy\cite{voigt2017eu, nayak2022real}. Federated learning (FL) is a paradigm that allows multiple institutions to collaboratively train a model while keeping raw data isolated \cite{yin2021comprehensive, fu2023client}. FedAvg~\cite{chellapandi2023federated} is the classic FL method where in each round, clients train local models on their private data and send only the resulting updates to a central server. The server aggregates these updates into a new global model and redistributes it to all clients. This process preserves data privacy by sharing only model parameters without exposing raw data.



When applying FL to multi-pulse MRI, its data heterogeneity, including the differences in tissue sensitivity across pulse, variability in scanner hardware and acquisition protocols, and diversity in patient populations results in suboptimal convergence and poor generalizability of the global model~\cite{sabah2024model}. To address this issue, some approaches extended FedAvg by introducing adaptive weights or regularization terms during aggregation to narrow the gap between local and global models~\cite{FedProx, ilhan2023scalefl}. Other works focus on enriching the diversity of training data in each client. For instance, ~\cite{xu2023bias, chen2023fraug} have employed data augmentation techniques to expand the number of samples each client.



However, the effect of data heterogeneity has not been significantly alleviated by using conventional FL~\cite{mendieta2022local}. 
To improve performance, researchers have introduced personalized federated learning (PFL), which comprises two principal strategies, global model personalization and personalized model learning~\cite{tan2022towards}. In the first strategy, each client applies targeted adaptations to the shared global model, allowing a single model to better accommodate the diverse distributions of all participants. The second strategy modifies the FL aggregation mechanism to train a distinct and highly generalizable model specifically for each client.

Apart from the effect of data heterogeneity, multi-pulse data are larger, which necessitates models with a large number of parameters to effectively fit. It significantly increases communication overhead in the context of FL, preventing the timely synchronization of global model updates, especially when there is a large number of clients~\cite{jiang2023complement}. Previous studies tackle the problem by compressing local updates using techniques such as quantization, pruning, and singular value decomposition (SVD)~\cite{ pfeiffer2023federated}. Alternatively, some methods adopt client selection strategies that prioritize clients with more representative data or better local performance~\cite{fraboni2021clustered, tang2022fedcor}. By reducing the volume of data sent to the server, these approaches improve the practicality of FL in bandwidth-constrained environments.

In this study, we propose a \textbf{C}ommunication-\textbf{E}fficient \textbf{Per}sonalized \textbf{Fed}erated Learning (CEPerFed) method for multi-pulse MRI classification. On the client side, we introduce two complementary gradients to guide the optimization of the local model, namely the historical risk gradient and the historical average gradient. During the transmission phase, we achieve efficient collaborative updates between the client and server by reducing communication costs. The main contributions of the paper are summarized as follows:

\begin{enumerate}
      \item We propose a novel PFL method CEPerFed that mitigates the effect of heterogeneity in multi-pulse MRI data and hierarchical SVD (HSVD) strategy to reduce the number of model parameters transmitted during training.
      \item To mitigate the effect of data heterogeneity, we use a risk matrix to weight gradients of other clients for each client, yielding a historical risk gradient that benefits local model optimization. Meanwhile, a historical average gradient computed across all clients both guides updates to the risk matrix and keeps local models aligned with the global optimization direction.
    \item The HSVD strategy categorizes convolutional layers into three distinct parts based on their functional characteristics and applies customized rank selection strategies to each. Before transmission, clients compress the model updates into low-rank factors and send them to the server.
    \item We design multi-pulse MRI combinations for comparative experiments and demonstrate that incorporating multi-pulse settings improves the disease classification performance of CEPerFed. To the best of our knowledge, this is the first systematic study of multi-pulse MRI classification. We also conduct ablation studies and parameter analyses that confirm the necessity of each module.
\end{enumerate}

\section{Related Work}

\subsection{Personalized Federated Learning}


The classic FedAvg~\cite{FedAvg} method assumed independent and identically distributed (IID) data across clients, while in real-world scenarios the data is non-IID (data heterogeneity), which significantly degrades its performance. To mitigate this issue, FedProx~\cite{FedProx} augmented each client’s local objective with a proximal term that penalized deviation from the global model. FedCD~\cite{FedCD} employed scores computed on a validation set to weight client contributions during aggregation. FedNP~\cite{FedNP} approximated the global data distribution via the product of a prior and a likelihood term. While these approaches are effective to some extent, these methods often lack generalizability under severe heterogeneity.

Personalization has also attracted considerable attention as a way to cope with data heterogeneity, leading to the development of the PFL paradigm. In this framework, the objective shifts from learning a single global model to training individualized models for each participating client. These tailored models are specifically optimized to perform effectively on the unique data distribution and task requirements of their respective clients. For example, inspired by FedAvg, PerFedAvg~\cite{fallah2020personalized} fine-tuned the global model on local data after each communication round. PGFed~\cite{PGFed} used based on Taylor expansion risk approximation to define the objective for each client, and FedALA~\cite{FedALA} adaptively defined local and global models during aggregation. Although existing PFL methods have made some progress, they primarily focus on local client optimization and often overlook the global collaborative knowledge shared. In contrast, our method employs coordinated local and global optimization is employed to mitigate the effects of data heterogeneity.

\subsection{Communication Optimization for Federated Learning}

In FL, clients need to frequently exchange data with the server to acquire global collaborative knowledge for optimizing their local models. MRI data from multiple institutions exhibit larger volume and more complex feature distributions, which require clients to train models with greater parameter counts. However, in scenarios with limited bandwidth or a large number of participating clients, transmitting such large models can incur significant communication overhead. This may congest client-server communication and cause some clients’ updates to lag, hindering global convergence. To alleviate this bottleneck, TurboSVM-FL~\cite{TurboSVM} used support vector machines to selectively aggregate class embeddings while incorporating max margin regularization to accelerate convergence, indirectly reducing communication costs. FedSLR~\cite{huang2023fusion} learnt a global shared representation via regularization and then applied SVD to extract its principal components.

Although existing SVD-based compression methods reduce parameter transmission, the decomposed matrices often still contain considerable redundant or even harmful information. To preserve model performance while curbing redundant transmission, we propose a HSVD strategy. HSVD leverages the structural characteristics of convolutional layers to partition them into three distinct categories, each of which employs a customized SVD strategy.

\begin{figure*}[htbp]  
	\centering
	\includegraphics[width=1\linewidth,scale=1.0]{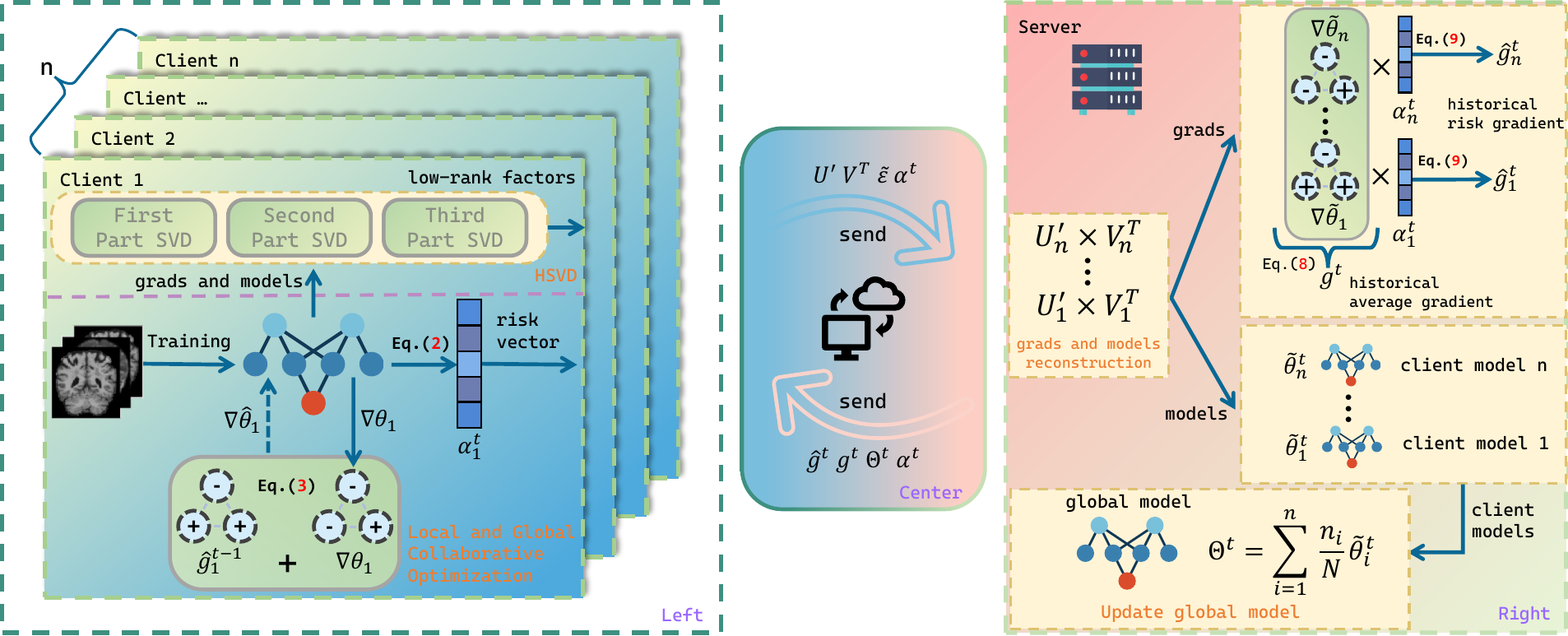}
    \caption{Overview of CEPerFed. Left: Each client updates its local model and historical risk gradient through local and global collaborative optimization module. After local training, clients decompose their gradients and parameters into low-rank factors using the proposed HSVD strategy and upload these factors to the server. Right: The received factors are reconstructed into each client’s gradients and models. The server then computes the historical average gradient and the historical risk gradient via Eq.~(\ref{eq12}) and Eq.~(\ref{eq13}), respectively, and updates the global model based on the aggregated parameters. Center: The server sends the updated gradients and the new global model to clients, which then proceed to the next training round.}
	\label{framework_img}
\end{figure*}

\subsection{Multi-pulse MRI}
This study innovatively designs a multi-pulse MRI experimental paradigm. By leveraging the complementary strengths of different pulse imaging, our method significantly improves disease classification performance and provides a new perspective for future MRI disease diagnosis research. To the best of our knowledge, this represents the first systematic effort targeting multi-pulse MRI classification, distinguishing our work from multimodal approaches discussed below.

Multimodal disease classification can significantly enhance the classification performance of model by integrating complementary information from different imaging modalities. When a modality fails to delineate a lesion clearly due to imaging noise, low resolution, or poor tissue contrast, other modalities often provide structural cues to compensate. MDL-Net~\cite{ronghe1} used a neural network to capture both local correlations and global dependencies between MRI and PET, thereby strengthening cross-modal feature representations. SGCN~\cite{ronghe3} constructed a graph from multimodal imaging and phenotypic data and employs a graph convolutional network to learn region and connection importance probabilities, enabling identification of related ROIs and pathology-specific brain network connections. ASMFS~\cite{adp1} designed feature similarities in a low-dimensional embedding space and select the most discriminative subset of features for training, further enhancing the fusion of multimodal information.

\section{Method}

\subsection{Overview}
During local model optimization, the local and global collaborative optimization module first corrects the raw batch gradient using the historical risk gradient according to Eq.~(\ref{eq2}) and updates the model parameters. Subsequently, the risk matrix is updated using Eq.~(\ref{eq1}). After finishing local training, each client applies the HSVD strategy to decompose both gradients and models, retaining only the most informative low-rank factors for transmission to the server.  At the server, the received low-rank factors are reassembled by matrix multiplication to form approximate client gradients and models. The reconstructed gradients are used to compute the historical average gradient and the historical risk gradient via Eq.~(\ref{eq12}) and Eq.~(\ref{eq13}), respectively. The global model is updated using the same aggregation rule as FedAvg. Finally, the server sends the computed gradients to the corresponding clients and broadcasts the updated global model for the next training round.

\subsection{Preliminaries}
\label{method_1}
\subsubsection{Personalize Federated Learning}
PFL extends FL by addressing the effect of data heterogeneity in client. In PFL, each client $i$ has a unique private dataset $D_i = \{(x_j, y_j)\}_{j=1}^{N_i}$, where $N_i$ denotes the size of the $D_i$. Coordinated by a server, PFL trains a local model $\theta_i$ on its private dataset $D_i$. The goal of PFL is to minimize a global objective in order to obtain local models with strong generalization. Formally, this can be expressed as:
\begin{equation}
\left \{ \theta_1, \theta_2, \dots, \theta_n \right \} = \arg\min \sum_{i=1}^{n} w_i \mathcal{L}_i(\theta_i),
\end{equation}
where $\theta_i$ represents the model parameters for client $i$. The weight $w_i$ is typically set as $w_i=\frac{N_i}{N}$, where $N$ is the total number of samples across all clients. $\mathcal{L}_i(\theta_i) = \mathbb{E}_{D_i \sim P_i}[f_i(\theta_i, D_i)]$ is the empirical risk for client $i$, which is represented by the loss function in our method. The training process in PFL involves two key stages:

\textbf{Client Training:} In the $t$-th round of local updates, client $i$ initializes the local model using the global model $\Theta^{\,t}$ received from server. It then performs iterative optimization using its local private dataset $D_i$, updating the local model parameters $\theta_i^{\,t}$. Finally, client $i$ sends its gradients $\nabla \theta_i^{\,t}$ and model parameters $\theta_i^{\,t}$ to the server.

\textbf{Collaborative Update:} During the collaborative update stage, clients transmit their models parameters and gradients to the server. The server aggregates these uploads to obtain updated global model parameters and gradients, which are then sent back to the corresponding clients.

\subsubsection{Singular Value Decomposition}
SVD is a matrix factorization method widely used for handling high-dimensional data. For a matrix \( A \in \mathbb{R}^{m \times n} \), it can be decomposed as \( A = U \Lambda V^\top \), where \( U \) and \( V \) are orthogonal matrices, and \( \Lambda \) is a diagonal matrix containing non-negative singular values \( \sigma_1 \geq \sigma_2 \geq \cdots \geq \sigma_R \geq 0 \), where $R = \text{rank}(A)$ represents the rank of the matrix. Furthermore, the magnitude of singular values reflects the importance of each component in the orthogonal matrices $U$ and $V$. Exploiting this SVD property, we approximate the original matrix by retaining the leading $r$ ($r \ll R$) singular values together with the corresponding top-$r$ submatrices of the orthogonal matrices $U$ and $V$.

SVD is applied to the matrices composed of each client’s model parameters \( \theta_i^t \) and corresponding gradients \( \nabla \theta_i^t \). Only the most informative submatrices are retained as low-rank factors and send to the server. Upon receiving these factors, the server reconstructs approximate model parameters and gradients by simple matrix multiplication.

\subsection{Local and Global Collaborative Optimization}
\label{method_2}

To mitigate the impact of data heterogeneity, each client has a historical risk gradient provided by the server, which is computed based on the client's risk vector and serves as a correction term. This term is combined with the raw gradient to produce a corrected gradient that guides the local model update. Specifically, the server maintains a historical risk matrix $\alpha$, where each client $i$ possesses its corresponding risk vector $\alpha_i$. The element $\alpha_{ij}^t$ in this vector quantifies the trust weight that client $i$ assigns to the gradients from client $j$ during local optimization. The risk vector is updated dynamically via:

\begin{equation}
\label{eq1}
\alpha_{ij}^t \leftarrow \max\left(\alpha_{ij}^{t-1} - \lambda \cdot (M_j + \langle \theta^{\,t}_i, g^{t-1} \rangle),\ 0\right),
\end{equation}
where \(\lambda\) controls the update step size and $\theta^{\,t}_i$ denotes model parameters of client $i$ in round $t$. $M_j=\mathcal{L}_{\mathrm{CE}}-\langle \nabla\theta_j,\,\theta_j \rangle$, where $\mathcal{L}_{\mathrm{CE}}$ denotes the cross-entropy loss, measures the directional alignment between current model gradient of client \(j\) and its parameters by the inner product of \(\nabla\theta_j\) and \(\theta_j\). A positive inner product $\langle \nabla\theta_j,\,\theta_j\rangle>0$ indicates an effective update and thus reduces $M_j$, while a misaligned or adverse update increases $M_j$. To ensure these local corrections remain consistent with global training dynamics, we incorporate an update direction consistency term $\langle \theta^{t}_i, g^{t-1}\rangle$ into the risk vector update. Here $g^{t-1}$ is the global historical average gradient computed as the weighted average of reconstructed client gradients (Eq.~(\ref{eq12}) in Sec.~\ref{sec3.D}). This term penalizes local optimizations that conflict with the global optimization direction, preventing personalized updates from diverging from the global training objective. Finally, the outer $\max(\cdot,0)$ enforces nonnegativity of the risk vector by truncating negative values to zero.

During client side optimization, the raw batch gradients are corrected before being used to update model parameters, as illustrated by the green box labeled Eq.~(\ref{eq2}) in the left parrt of Fig.~\ref{framework_img}. Specifically, in round $t$, client $i$ incorporates the historical risk gradient $\hat{g}_{i}^{\,t-1}$, which was computed by the server in the previous round using the risk vector $\alpha^{\,t-1}_i$ from Eq.~(\ref{eq13}) in Sec.~\ref{sec3.D}), into its current gradient $\nabla\theta_{i}$ to form the corrected gradients. Formally, this update is written as:
\begin{equation}
\label{eq2}
\quad \nabla\hat{\theta}_{i} \leftarrow \nabla\theta_{i} + \hat{g}_{i}^{\,t-1}.
\end{equation}

Incorporating the historical risk gradient enables each client to account for the reliability of other clients’ updates during local training. Importantly, both the historical risk gradients and the historical average gradient are computed centrally on the server and distributed to the corresponding clients. Because clients only receive these aggregated gradients and cannot independently reconstruct other clients’ raw gradients, data privacy is effectively preserved throughout the collaborative learning process.

\begin{figure}[htbp]  
	\centering
	\includegraphics[width=\linewidth,scale=1.0]{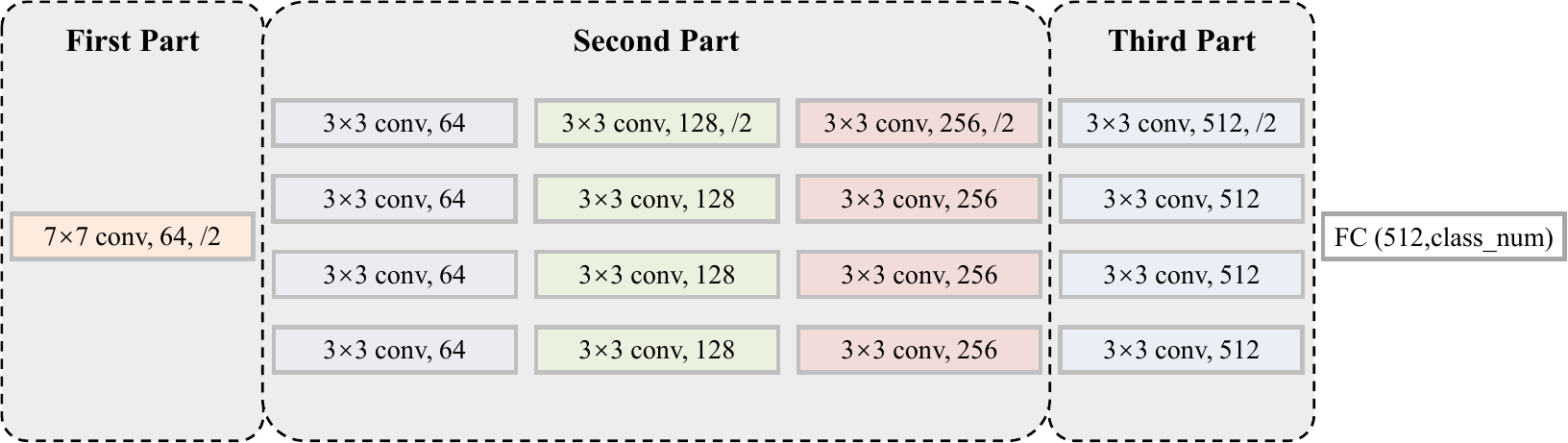}
    \caption{Hierarchical partitioning of ResNet-18 into three distinct parts for the HSVD strategy.}
	\label{layer_img}
\end{figure}

\subsection{Hierarchical Singular Value Decomposition}

\begin{figure*}[htbp]  
	\centering
    \subfigure[Residual compensated SVD for first part.]
    {\includegraphics[width=0.329\linewidth]{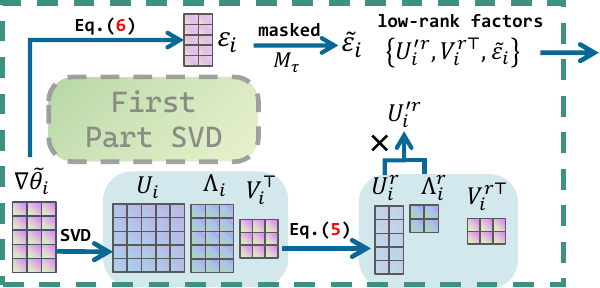}}
    \subfigure[Dynamic rank selection for the second part.]
    {\includegraphics[width=0.329\linewidth]{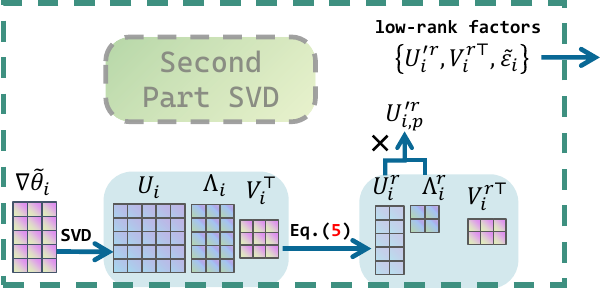}}
    \subfigure[Group-based fixed-rank decomposition for the third part.]
    {\includegraphics[width=0.329\linewidth]{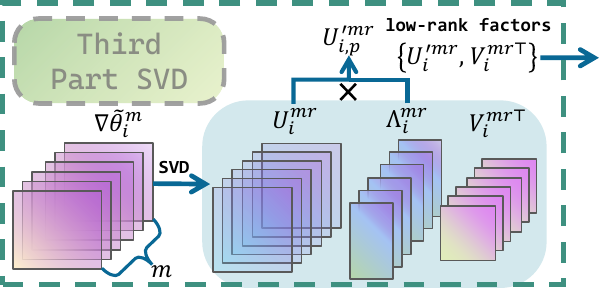}}
	\caption{Execution workflow of the proposed HSVD strategy across different network parts. The right arrows indicate the transmission direction of the compressed low-rank factors and do not imply a sequential dependency between the parts.}
	\label{fenceng_img}
\end{figure*}

\label{h-svd}

After completing the above training, clients designed an HSVD strategy to reduce the parameters passed to the server, which compresses model gradients and parameters using three distinct methods. The proposed HSVD is designed based on ResNet-18, whose convolutional layers are partitioned according to the structure illustrated in Fig.~\ref{layer_img}. The same compression strategy is applied to both the model parameters and their corresponding gradients. For clarity and brevity, only the gradient decomposition procedure is described in the following sections.

\subsubsection{Residual Compensated SVD for First Part}
\label{sec:3.C}
In convolutional neural networks, the first convolutional part is responsible for extracting low-level features e.g., edges and textures, the quality of these features significantly affects subsequent feature processing. Therefore, applying a SVD directly to the gradients of this part can discard optimized information. To mitigate this issue, we propose a residual compensated gradient decomposition (Fig.~\ref{fenceng_img}(a)) for the first part of the convolutional layer, which is designed to preserve more of those important gradients details during SVD.

Concretely, we reshape the client's four-dimensional gradient matrix $\nabla\Bar{\theta}_{i}\in\mathbb{R}^{c_o\times c_i\times k_h\times k_w}$ into a two-dimensional $\nabla\tilde{\theta}_{i}\in\mathbb{R}^{c_o\times a}$, where $a = c_i\cdot k_h\cdot k_w$. Here $c_o$ and $c_i$ denote the numbers of output and input channels, respectively, while $k_h$ and $k_w$ denote the convolution kernel height and width. We apply SVD to the reshaped matrix to obtain the three matrices:
\begin{equation}
\label{eq3}
\nabla\tilde{\theta}_{i} \;=\; U_{i}\,\Lambda_{i}\,V_{i}^\top\;,
\end{equation}
where $U_{i}\in\mathbb{R}^{c_o\times c_o}$ is the left singular matrix, $\Lambda_{i}\in\mathbb{R}^{c_o\times a}$ is a diagonal matrix composed of singular values $\sigma_1,\sigma_2,\dots,\sigma_n$, and $V_{i}^\top\in\mathbb{R}^{a\times a}$ is the transpose of the right singular matrix. 

The essence of SVD is to decompose a complex linear transformation into three successive operations: rotation by $U_{i}$, scaling by $\Lambda_{i}$, and rotation by $V_{i}$. The diagonal entries of $\Lambda_{i}$, i.e. the singular values sorted in descending order, measure the importance of the corresponding orthogonal basis vectors (the columns of $U_{i}$ and $V_{i}$). Consequently, the diagonal singular value matrix is the core component of the decomposition. Leveraging this property, we determine the retained rank \(r\) based on a matrix energy criterion:
\begin{equation}
\label{eq4}
r \;=\;\min\Bigl\{k\in\mathbb{Z}^+\Bigm|\frac{\sum_{i=1}^k\sigma_i^2}{\sum_{i=1}^{\min(a,b)}\sigma_i^2}\geq\eta\Bigr\}.
\end{equation}
Here the matrix energy is defined as the sum of squared singular values after SVD, which reflects the information richness of the matrix. The numerator $\sum_{i=1}^{k}\sigma_i^2$ denotes the energy contained by the first $k$ singular values, while the denominator $\sum_{i=1}^{\min(a,b)}\sigma_i^2$ gives the total energy across all singular values $\sigma_i$. By setting the threshold $\eta=0.9$, we ensure that at least 90\% of the gradient information is preserved within the selected rank $r$. Because the retained rank $r$ is computed dynamically via Eq.~(\ref{eq3}), we refer to this method as dynamic rank selection. For simplify computation, we similarly retain only the top \( r \) columns of \(U_{i}\) and \(V_{i}\), using this same rank \( r \) as determined above.

The dynamic rank selection method preserves most of the information in the gradient matrix, but some gradients may still lose substantial detail. For those gradients we introduce a sparse residual compensation term. Specifically, after determining the retained rank $r$ and obtaining the low-rank factors $\{U_{i}^r,\,\Lambda_{i}^r,\,V_{i}^{r\top}\}$, the client calculates the residual of the gradient \(\nabla\tilde{\theta}_{i}\) as follows:
\begin{equation}
\label{eq5}
\varepsilon_{i}
= \nabla\tilde{\theta}_{i}
- U_{i}^r\,\Lambda_{i}^r\,V_{i}^{\,r\top}\;.
\end{equation}

Next, to selectively recover the components most truncated by the low-rank decomposition, we first sort the absolute values of $\varepsilon_{i}$. Based on this ordering we construct a binary mask $M_{\tau}$ that sets to one the top 10\% largest entries (i.e., the positions that lost the most information) and to zero all other entries. The mask is applied element-wise to the residual $\varepsilon_{i}$, and the result is scaled by a compensation factor $\gamma$ to produce the final compensated residual
$\tilde{\varepsilon}_{i} = \gamma \,\varepsilon_{i} \odot M_{\tau}$. This procedure restores the most severely truncated gradient components of the first part while keeping the overall amount of transmitted information small.

Finally, the client sends only the low-rank factors \(\{U_{i}^{\prime r},\,V_{i}^{r\top},\,\tilde{\varepsilon}_{i}\}\), where \(U_{i}^{\prime r} = U_{i}^r\,\Lambda_{i}^r\). The server reconstructs the approximate first part gradients as $\nabla\tilde{\theta}_{i} \approx U_{i}^{\prime r}V_{i}^{r\top} + \tilde{\varepsilon}_{i}$.

\subsubsection{Dynamic Rank Selection for Second Part}
\label{hsvd-2}

Fitting the more complex features in the second convolutional layer demands a higher number of parameters, consequently exhibiting greater parameter redundancy than in the first part. Given this redundancy, applying residual compensation would impose unnecessary computational and communication overhead. Guided by the methodology in Sec.~\ref{sec:3.C}, we omit the computation of the residual compensation term. The overall procedure is illustrated in the middle panel of Fig.~\ref{fenceng_img}(b).

We reshape the middle part gradient into a two-dimensional matrix and apply the same matrix energy criterion to choose a retained rank $r$ so that at least 90\% of the gradient information is preserved. The client then retains the corresponding low-rank factors and sends these factors $\{U_{i}^{\prime r},\,V_{i}^{r\top}\}$ to the server. Upon reception, the server recovers an approximate gradient by simple matrix multiplication $ \nabla\tilde{\theta}_{i}\approx U_{i}^{\prime r}V_{i}^{r\top}$, greatly reducing transmitted data while preserving the dominant gradient.

\subsubsection{Group-Based Fixed-Rank Decomposition for Third Part}

Deeper convolutional layers extract features that transition from low-level edges and textures to high-level semantic representations, which typically comprise global abstract information and exhibit strong inter channel correlations~\cite{yamanakkanavar2022mf2, SHU2024102800}. However, applying a unified SVD to the entire gradient matrix would disrupt these critical correlations. Moreover, the substantial parameter volume in these layers makes global, energy-based rank selection computationally expensive. To address these limitations, we propose a group-based fixed-rank SVD strategy. This approach first partitions the gradient matrix of the third part along the output channel into several smaller groups. Each group then undergoes independent fixed-rank SVD, retaining only the top-$r$ components to form low-rank factors. The overall procedure is illustrated in Fig.~\ref{fenceng_img}(c).


Specifically, for the gradient matrix \(\nabla\Bar\theta_{i}\in\mathbb{R}^{c_o\times c_i\times k_h\times k_w}\) of the third part, we partition it along the output channel \(c_o\) into \(M = c_o/c\) equally sized groups. This results in a set of sub-matrices \(\{\nabla\bar{\theta}_i^{\,m}\}_{m=1}^M\), where each group \(\nabla\bar{\theta}_i^{\,m}\) contains \(c\) channels. These groups are then reshaped into $\nabla\tilde{\theta}_{i}^{\,m}\in\mathbb{R}^{c\times a}$ and decomposed via a fixed-rank \(r\) SVD, yielding \(\{U_{i}^{\,m r},\Lambda_{i}^{\,m r},V_{i}^{\,m r\top}\}\) for the $m$-th group. To obtain the balance between information preservation and communication efficiency, we fix the group size \(c=64\) and the rank \(r=16\). Following the procedure in Sec.~\ref{hsvd-2}, the client sends only the low-rank factors \(\{U_{i}^{\prime m r},\,V_{i}^{\,m r\top}\}\), where \(U_{i}^{\prime m r} = U_{i}^{m r}\,\Lambda_{i}^{m r}\). Upon receipt, the server first approximates the original gradient of each group via the matrix  multiplication, and then concatenates these reconstructed groups along the output channels:
\begin{equation}
\nabla\tilde{\theta}_{i}
= \bigoplus_{m=1}^{M}\bigl(U_{i}^{\prime m r}\,V_{i}^{\,m r\top}\bigr).
\end{equation}

Given that fully connected layer has relatively few parameters yet is highly sensitive to decision boundaries, we transmit its gradients without compression to preserve the integrity of the final classification output.

\subsection{Aggregation and Gradient Construction}
\label{sec3.D}

In each round of local training, clients apply HSVD strategy to compress both their model parameters and gradients, sending the resulting low‑rank factors only. The server reconstructs the gradient $\nabla\Bar\theta_{i}$ of each client into $\nabla\tilde{\theta}_{i}$ using matrix multiplication based on the received low-rank factors. Then, these reconstructed gradients are used to calculate the global historical average gradient $g^t$ and each client's historical risk gradient $\hat{g}_i^t$. The updated global model and gradients are then sent back to the clients for the next round of local optimization.

Specifically, the server sums the reconstructed gradients $\nabla\tilde{\theta}_{i}$ of \( n \) clients and then scales them to obtain the global historical average gradient, expressed as follows:
\begin{equation}
\label{eq12}
g^t = \frac{\delta}{n} \sum_{i=1}^{n} \nabla\tilde{\theta}^{\,t}_i,
\end{equation}
where the hyperparameter \(\delta\) acts as a scaling factor to suppress excessive perturbations of the risk vector \(\alpha^t_i\) induced by the global gradient direction consistency constraint (Eq.~\ref{eq1} for \(\langle \Theta^t, g^{t-1}\rangle\)). The historical average gradient integrates the gradient information of all clients and provides a reference for the global consistency update of the local model.

Next, the server uses each client’s risk vector \(\alpha_i^t\) to perform a weighted combination of all clients reconstructed gradients, to construct the historical risk gradient for client \(i\):
\begin{equation}
\label{eq13}
\hat{g}_i^t = \sum_{j=1}^{n} \alpha^t_{ij} \nabla\tilde{\theta}^{\,t}_i.
\end{equation}
The historical risk gradient highlights collaborative knowledge beneficial to local client $i$ through the risk vector $\alpha^t_{i}$.

The server updates the global model via the aggregation rule $\Theta^{t}=\sum_{i=1}^n \frac{n_i}{N}\,\tilde{\theta}^{\,t}_i$. Subsequently, server distributes the updated parameters \(\Theta^t\), the historical average gradient \(g^t\), and the client-specific historical risk gradients \(\{\hat{g}_i^t\}_{i=1}^n\) to the clients for the next round of local training.


\section{Experiments}

\subsection{Experimental Setup}

\subsubsection{Multi-Pulse MRI}
To investigate the impact of multi-pulse MRI on disease classification, we employ three pulse sequences for our experiments: Gradient-recalled echo (\textbf{G}), Sagittal inversion recovery (\textbf{S}), and Gradient-recalled echo repeat (\textbf{R}). Their imaging characteristics are described as follows:

\begin{itemize}
    \item \textbf{G}: This pulse setting is widely used for brain and craniofacial imaging. It conducts magnetization preparation, rapid gradient echo acquisition, and reconstruction filtering to produce high quality structural brain images within a relatively short scan time.
    \item \textbf{S}: This pulse setting acquires images in the sagittal plane and is commonly employed for T1-weighted imaging. This imaging setup enhances contrast between different tissue types, particularly between gray and white matter in the brain.    
    \item \textbf{R}: This pulse sequence employs \textbf{G} resampling on targeted regions or subjects, thereby enhancing the signal-to-noise ratio while mitigating motion-induced artifacts.
\end{itemize}

To ensure input consistency across the different pulse imaging, all three-dimensional brain MRI scans were preprocessed using the FSL toolkit \cite{jenkinson2012fsl}, including image registration, skull stripping, and intensity normalization. Next, we extracted three axial slices along the Z-axis for each subject, namely the central slice and the slices five layers above and below the center, and stacked them as channels to form a single three-channel image for subsequent model input. After preprocessing the multiple follow-up data from each subject, 15,686 three-channel MRI images were obtained for model training. Their distribution by classification label is: CN (4,582), EMCI (2,840), LMCI (1,256), MCI (4,799), and AD (2,209).

\begin{table}
\centering
\caption{The number of samples for different pulse combinations.}
\label{tab:sequence_distribution}
\scriptsize
\begin{tabular}{@{}ccccccc@{}}  
\toprule
\textbf{Pulse} & \textbf{CN} & \textbf{EMCI} & \textbf{LMCI} & \textbf{MCI} & \textbf{AD} & \textbf{Total} \\
\midrule
\textbf{G}   & 2,539 & 2,053 & 955  & 2,200 & 1,146 & 8,893   \\
\textbf{S}   & 1,012 & 787  & 301  & 795  & 523  & 3,418   \\
\textbf{R}   & 1,031 & {0}  & {0}  & 1,804 & 540  & 3,375   \\
\textbf{GS}  & 3,551 & 2,840 & 1,256 & 2,995 & 1,669 & 12,311  \\
\textbf{GR}  & 3,570 & 2,053 & 955  & 4,004 & 1,686 & 12,268  \\
\textbf{GSR} & 4,582 & 2,840 & 1,256 & 4,799 & 2,209 & 15,686  \\
\bottomrule
\end{tabular}
\end{table}

\begin{table*}[htb]
  \centering
  \caption{Accuracy performance evaluation under different pulse combinations and non-IID conditions (Client-wise Mean ± Standard Deviation; Best in \textbf{bold} and second with \underline{underlined}).}
  \label{tab:merged}
  \footnotesize
  \sisetup{
    table-number-alignment=center,
    separate-uncertainty,
    table-format=2.2(2),
    uncertainty-separator={\,\scriptsize}
  }
  \begin{tabular}{@{}l c *{6}{c}@{}}
    \toprule
\multirow{2}{*}{\small\textbf{Method}} & \raisebox{0.6ex}{\textbf{T\textsubscript{1}}} & \raisebox{0.6ex}{\textbf{T\textsubscript{2}}} & \raisebox{0.6ex}{\textbf{T\textsubscript{3}}} 
& \raisebox{0.6ex}{\textbf{T\textsubscript{4}}} & \raisebox{0.6ex}{\textbf{T\textsubscript{5}}} & \multirow{2}{*}{\small\textbf{Avg.}} \\
& \footnotesize CN vs. MCI & \footnotesize MCI vs. AD & \footnotesize CN vs. AD 
& \footnotesize CN vs. MCI vs. AD & \footnotesize CN vs. EMCI vs. LMCI vs. AD & \\
\midrule

    \multicolumn{7}{c}{\textbf{G pulse}} \\
    \midrule
    FedAvg \cite{FedAvg}        & 72.36$\pm$5.84  & 66.87$\pm$8.06 & 79.21$\pm$5.63 & 62.07$\pm$4.37 & 67.46$\pm$6.79  & 69.59$\pm$6.14 \\
    FedProx \cite{FedProx}       & 72.77$\pm$5.77  & 72.06$\pm$4.55 & 74.93$\pm$7.76 & 60.78$\pm$4.96 & 67.44$\pm$6.65  & 69.60$\pm$5.94 \\
    FedCD \cite{FedCD}         & 71.65$\pm$5.03  & 72.33$\pm$5.71 & 79.43$\pm$5.73 & 59.75$\pm$5.79 & 64.06$\pm$6.88  & 69.44$\pm$5.83 \\
    FedNP \cite{FedNP}         & \underline{75.30$\pm$2.08}  & \underline{76.50$\pm$3.01} & 81.74$\pm$3.34 & \underline{69.53$\pm$2.44} & \underline{74.12$\pm$2.55}  & \underline{75.44$\pm$2.68} \\
    FedALA \cite{FedALA}        & 74.46$\pm$3.43  & 75.91$\pm$2.79 & \underline{81.96$\pm$3.39} & 66.06$\pm$2.95 & 71.53$\pm$4.95  & 73.98$\pm$3.50 \\
    PGFed \cite{PGFed}         & 72.35$\pm$6.08  & 72.70$\pm$6.82 & 81.22$\pm$4.61 & 61.41$\pm$5.14 & 67.80$\pm$4.18  & 71.10$\pm$5.37 \\
    TurboSVM-FL \cite{TurboSVM}      & 72.10$\pm$5.66  & 69.00$\pm$6.97 & 77.78$\pm$5.66 & 62.18$\pm$4.98 & 61.55$\pm$7.03  & 68.52$\pm$6.06 \\
    FedDecorr \cite{FedDecorr}  & 74.68$\pm$3.34  & 74.04$\pm$4.54 & 81.91$\pm$4.18 & 65.38$\pm$3.33 & 70.97$\pm$4.97  & 73.40$\pm$4.07 \\
    FedAu \cite{FedAu}         & 74.23$\pm$4.56  & 69.45$\pm$7.86 & 76.65$\pm$8.62 & 62.59$\pm$5.03 & 65.15$\pm$7.99  & 69.61$\pm$6.81 \\
    \midrule
    \rowcolor{gray!10}
    CEPerFed    & \textbf{78.12$\pm$3.46}  & \textbf{77.30$\pm$2.13} & \textbf{84.49$\pm$2.70} & \textbf{69.73$\pm$2.26} & \textbf{77.35$\pm$2.01}  & \textbf{77.40$\pm$2.51} \\
    \midrule

    \multicolumn{7}{c}{\textbf{GS pulse combination}} \\
    \midrule
    FedAvg \cite{FedAvg}        & 72.96$\pm$8.17  & 76.68$\pm$6.37 & 78.89$\pm$6.57 & 66.31$\pm$4.71 & 64.81$\pm$5.81  & 71.93$\pm$6.33 \\
    FedProx \cite{FedProx}       & 73.16$\pm$6.25  & 73.40$\pm$8.25 & 76.80$\pm$5.39 & 63.32$\pm$7.56 & 65.37$\pm$7.74  & 70.41$\pm$7.04 \\
    FedCD \cite{FedCD}         & 72.77$\pm$7.02  & 72.87$\pm$5.75 & 78.93$\pm$4.72 & 63.41$\pm$6.76 & 62.89$\pm$6.88  & 70.17$\pm$6.23 \\
    FedNP \cite{FedNP}         & \underline{80.89$\pm$2.42}  & \underline{80.66$\pm$2.34} & 83.10$\pm$3.16 & \textbf{75.92$\pm$3.35} & 76.98$\pm$2.10  & \underline{79.51$\pm$2.67} \\
    FedALA \cite{FedALA}        & 79.05$\pm$3.65  & 79.19$\pm$3.33 & 83.58$\pm$2.77 & 71.12$\pm$4.41 & \underline{79.05$\pm$3.65}  & 78.40$\pm$3.56 \\
    PGFed \cite{PGFed}         & 72.37$\pm$6.66  & 75.53$\pm$6.86 & 77.05$\pm$6.43 & 63.39$\pm$7.59 & 66.84$\pm$7.55  & 71.04$\pm$7.02 \\
    TurboSVM-FL \cite{TurboSVM}      & 72.26$\pm$7.54  & 74.46$\pm$7.13 & 76.36$\pm$5.85 & 62.44$\pm$8.02 & 64.30$\pm$7.67  & 69.96$\pm$7.24 \\
    FedDecorr \cite{FedDecorr}  & 78.71$\pm$3.56  & 80.58$\pm$3.17 & \underline{85.49$\pm$2.18} & 70.17$\pm$5.30 & 73.30$\pm$3.39  & 77.65$\pm$3.52 \\
    FedAu \cite{FedAu}         & 71.57$\pm$7.28  & 75.96$\pm$7.10 & 78.94$\pm$5.16 & 67.38$\pm$5.28 & 67.84$\pm$6.21  & 72.34$\pm$6.21 \\
    \midrule
    \rowcolor{gray!10}
    CEPerFed    & \textbf{83.61$\pm$1.87}  & \textbf{82.50$\pm$1.23} & \textbf{85.98$\pm$1.17} & \underline{75.91$\pm$1.53} & \textbf{79.06$\pm$1.55}  & \textbf{81.41$\pm$1.47} \\
    \midrule

    \multicolumn{7}{c}{\textbf{GR pulse combination}} \\
    \midrule
    FedAvg \cite{FedAvg}        & 79.41$\pm$7.05  & 75.08$\pm$6.09 & 85.02$\pm$7.04 & 71.39$\pm$6.77 & 73.05$\pm$4.77  & 76.79$\pm$6.34 \\
    FedProx \cite{FedProx}       & 82.07$\pm$6.43  & 77.36$\pm$6.09 & 84.35$\pm$6.12 & 71.70$\pm$7.84 & 73.20$\pm$7.06  & 77.74$\pm$6.71 \\
    FedCD \cite{FedCD}         & 79.04$\pm$6.71  & 74.87$\pm$6.51 & 78.92$\pm$6.41 & 71.91$\pm$6.30 & 73.20$\pm$7.06  & 75.59$\pm$6.60 \\
    FedNP \cite{FedNP}         & \textbf{88.26$\pm$1.40}  & 82.52$\pm$3.98 & 88.93$\pm$2.94 & \underline{83.33$\pm$2.30} & 71.38$\pm$5.42  & \underline{82.88$\pm$3.21} \\
    FedALA \cite{FedALA}        & 85.36$\pm$4.09  & 77.52$\pm$6.37 & 87.39$\pm$3.70 & 78.97$\pm$4.28 & \underline{79.44$\pm$3.20}  & 81.74$\pm$4.33 \\
    PGFed \cite{PGFed}         & 80.75$\pm$5.34  & 78.39$\pm$8.38 & 84.62$\pm$5.43 & 76.46$\pm$6.31 & 75.75$\pm$5.40  & 79.19$\pm$6.17 \\
    TurboSVM-FL \cite{TurboSVM}      & 77.30$\pm$8.43  & 74.62$\pm$4.97 & 82.93$\pm$6.35 & 71.46$\pm$8.04 & 71.79$\pm$5.86  & 75.62$\pm$6.73 \\
    FedDecorr \cite{FedDecorr}  & 83.58$\pm$4.48  & \underline{84.33$\pm$3.83} & \textbf{89.72$\pm$2.42} & 77.73$\pm$5.98 & 76.86$\pm$4.84  & 82.44$\pm$4.31 \\
    FedAu \cite{FedAu}         & 82.68$\pm$5.64  & 78.48$\pm$8.67 & 85.34$\pm$5.71 & 76.30$\pm$6.83 & 75.30$\pm$6.53  & 79.62$\pm$6.68 \\
    \midrule
    \rowcolor{gray!10}
    CEPerFed    & \underline{88.03$\pm$1.73}  & \textbf{86.61$\pm$2.77} & \underline{89.54$\pm$1.25} & \textbf{83.57$\pm$1.44} & \textbf{83.76$\pm$2.06}  & \textbf{86.30$\pm$1.85} \\
    \midrule

    \multicolumn{7}{c}{\textbf{GSR pulse combination}} \\
    \midrule
    FedAvg \cite{FedAvg}        & 83.64$\pm$5.42  & 82.54$\pm$5.16 & 85.20$\pm$5.78 & 72.32$\pm$8.36 & 72.43$\pm$5.90  & 79.23$\pm$6.12 \\
    FedProx \cite{FedProx}       & 83.32$\pm$5.39  & 81.26$\pm$5.59 & 83.70$\pm$7.94 & 69.41$\pm$9.28 & 73.34$\pm$5.53  & 78.21$\pm$6.75 \\
    FedCD \cite{FedCD}         & 82.35$\pm$4.87  & 81.06$\pm$6.81 & 82.86$\pm$7.59 & 71.73$\pm$7.17 & 70.60$\pm$6.93  & 77.52$\pm$6.67 \\
    FedNP \cite{FedNP}         & \underline{88.19$\pm$1.87}  & \underline{87.55$\pm$2.10} & \underline{90.37$\pm$1.31} & \underline{85.00$\pm$1.41} & 82.65$\pm$1.42  & \underline{86.75$\pm$1.62} \\
    FedALA \cite{FedALA}        & 84.61$\pm$3.11  & 85.68$\pm$5.77 & 88.77$\pm$3.28 & 72.49$\pm$10.97 & \textbf{84.61$\pm$3.11}  & 83.23$\pm$5.25 \\
    PGFed \cite{PGFed}         & 84.92$\pm$4.11  & 83.67$\pm$8.14 & 86.35$\pm$5.92 & 72.42$\pm$9.56 & 72.64$\pm$5.95  & 80.00$\pm$6.74 \\
    TurboSVM \cite{TurboSVM}      & 81.16$\pm$6.09  & 80.75$\pm$7.34 & 82.85$\pm$7.50 & 72.99$\pm$8.53 & 71.17$\pm$7.84  & 77.78$\pm$7.46 \\
    FedDecorr-FL \cite{FedDecorr}  & 84.54$\pm$3.74  & 86.51$\pm$4.95 & 89.65$\pm$3.13 & 79.17$\pm$5.51 & 77.87$\pm$3.45  & 83.55$\pm$4.16 \\
    FedAu \cite{FedAu}         & 82.88$\pm$5.60  & 81.25$\pm$6.20 & 83.80$\pm$6.98 & 70.92$\pm$9.85 & 74.44$\pm$5.15  & 78.66$\pm$6.76 \\
    \midrule
    \rowcolor{gray!10}
    CEPerFed    & \textbf{89.88$\pm$1.24}  & \textbf{90.99$\pm$1.61} & \textbf{92.26$\pm$1.66} & \textbf{86.74$\pm$1.39} & \underline{84.25$\pm$0.90}  & \textbf{88.82$\pm$1.36} \\
    \midrule

    \multicolumn{7}{c}{\textbf{GSR pulse combination under non-IID}} \\
    \midrule
    FedAvg \cite{FedAvg}        & \underline{68.69$\pm$7.29}  & 68.37$\pm$9.91  & 65.56$\pm$6.28  & 49.94$\pm$3.29 & 56.80$\pm$8.00   & 61.87$\pm$6.95 \\
    FedProx \cite{FedProx}       & 67.73$\pm$5.94  & 64.96$\pm$13.57 & 65.76$\pm$7.85  & 52.70$\pm$3.65 & 55.67$\pm$7.78   & 61.36$\pm$7.76 \\
    FedCD \cite{FedCD}         & 62.03$\pm$6.42  & 65.17$\pm$13.55 & 65.38$\pm$6.59  & 47.00$\pm$5.24 & 50.52$\pm$5.38   & 58.02$\pm$7.44 \\
    FedNP \cite{FedNP}         & 62.03$\pm$6.42  & 65.17$\pm$13.55 & 65.38$\pm$6.59  & 48.81$\pm$7.76 & 50.52$\pm$5.38   & 58.38$\pm$7.94 \\
    FedALA \cite{FedALA}        & 68.53$\pm$6.90  & \underline{69.77$\pm$4.61}  & 63.65$\pm$7.63  & 46.13$\pm$7.76 & \underline{56.95$\pm$5.23}   & 61.01$\pm$6.43 \\
    PGFed \cite{PGFed}         & 63.82$\pm$9.22  & 68.68$\pm$5.22  & 64.87$\pm$4.09  & 50.76$\pm$4.04 & 53.04$\pm$6.95   & 60.23$\pm$5.90 \\
    TurboSVM-FL \cite{TurboSVM}      & 67.93$\pm$6.24  & 67.96$\pm$8.43  & 65.57$\pm$7.86  & 49.11$\pm$4.53 & 54.65$\pm$8.22   & 61.04$\pm$7.06 \\
    FedDecorr \cite{FedDecorr}  & 65.02$\pm$10.21 & 65.48$\pm$15.83 & 62.52$\pm$11.67 & 46.74$\pm$6.55 & 46.29$\pm$18.92  & 57.21$\pm$12.64 \\
    FedAu \cite{FedAu}         & 68.46$\pm$6.82  & 71.07$\pm$8.32  & 66.41$\pm$7.29  & \underline{53.74$\pm$5.56} & 55.24$\pm$8.03   & \underline{62.98$\pm$7.20} \\
    \midrule
    \rowcolor{gray!10}
    CEPerFed    & \textbf{72.15$\pm$1.75}  & \textbf{78.06$\pm$2.36}  & \textbf{71.65$\pm$3.38}  & \textbf{61.86$\pm$1.96} & \textbf{66.59$\pm$0.86}   & \textbf{70.06$\pm$2.06} \\
    \bottomrule
  \end{tabular}
\end{table*}

\subsubsection{Task Design}
To systematically evaluate the effect of multi-pulse MRI on classification performance, we conducted experiments using the \textbf{G} pulse sequence as the baseline. The \textbf{G} was chosen because it is fast to acquire, provides high resolution, and is widely adopted in both clinical and research practice. Based on this baseline, we considered four pulse combinations: \textbf{G}, \textbf{GS}, \textbf{GR}, and \textbf{GSR}. Here, \textbf{S} supplies additional tissue-contrast information beyond \textbf{G}, while \textbf{R} helps mitigate motion-related artifacts. The number of samples for each combination is shown in Table~\ref{tab:sequence_distribution}.

For each pulse combination, we evaluate five classification tasks to cover different diagnostic scenarios: CN vs. MCI (\textbf{T\textsubscript{1}}); MCI vs. AD (\textbf{T\textsubscript{2}}); CN vs. AD (\textbf{T\textsubscript{3}}); CN vs. MCI vs. AD (\textbf{T\textsubscript{4}}); CN vs. EMCI vs. LMCI vs. AD (\textbf{T\textsubscript{5}}).

\subsubsection{Implementation Details}
All comparative methods were implemented using a ResNet-18 architecture and optimized with the Adam optimizer for a fair comparison. We fixed the batch size at 128, set the initial learning rate to $1\times10^{-4}$, and simulated five client nodes in each experiment. The dataset was randomly split into 80\% for training and 20\% for testing, with early stopping applied using a patience of 10 epochs to prevent overfitting. For the proposed HSVD strategy, the reduction step size $\lambda$ of the risk matrix $\alpha$ was set to $0.01$, and the global historical average gradient scaling factor $\delta$ was set to $0.1$.

\subsection{Performance Evaluation}

\subsubsection{Compared Methods}
To comprehensively evaluate CEPerFed on multi-pulse MRI disease classification, we compared it with four types of FL methods: (1) the baseline method FedAvg~\cite{FedAvg}; (2) methods designed to mitigate data heterogeneity including FedProx~\cite{FedProx}, FedCD~\cite{FedCD}, and FedNP~\cite{FedNP}; (3) PFL approaches FedALA~\cite{FedALA} and PGFed~\cite{PGFed}; (4) communication optimization methods, including TurboSVM-FL~\cite{TurboSVM}, FedDecorr~\cite{FedDecorr}, and FedAu~\cite{FedAu}.

Our experimental results are summarized in Table~\ref{tab:merged}. The first four subtables present the results for the four pulse combinations (\textbf{G}, \textbf{GS}, \textbf{GR}, and \textbf{GSR}), while the final subtable shows the performance of the \textbf{GSR} combination under non-IID settings.

\subsubsection{Analysis of \textbf{G} Results}
Our method achieves a mean accuracy of 77.40\% with a standard deviation of $\pm$2.51, representing a 7.81\% average improvement over FedAvg. FedCD and TurboSVM-FL underperform FedAvg, because they are optimized for natural image data at expense of generality. FedAvg is designed as a general purpose FL method typically yields robust performance across diverse data. Interestingly, the four-class task \textbf{T\textsubscript{5}} achieves higher accuracy than the three-class task \textbf{T\textsubscript{4}}. This is because the EMCI and LMCI classes are relatively easy to distinguish~\cite{arpanahi2025mapping}.

\subsubsection{Analysis of \textbf{GS} Results}
The \textbf{S} pulse sequence provides additional imaging perspectives and tissue contrast that complement and enrich the basic \textbf{G} pulse. Our method achieved an average accuracy of 81.41\%, a 4.01\% improvement over that uses \textbf{G}. Notably, the largest gain was observed in task \textbf{T\textsubscript{4}}, which improved by 6.18\%. This improvement is mainly driven by substantial performance increases on the more challenging tasks \textbf{T\textsubscript{1}} and \textbf{T\textsubscript{2}}, which therefore produce larger relative gains than \textbf{T\textsubscript{3}}. Other methods improved under the \textbf{GS} pulse combination, corroborating that additional imaging perspective and enhanced tissue contrast materially boost classification performance.

\subsubsection{Analysis of \textbf{GR} Results}
Under this setting, CEPerFed achieved an average accuracy of 86.30\%, an 8.9\% improvement over using \textbf{G} pulse alone. The most substantial gain was observed in task \textbf{T\textsubscript{4}}, which increased by 13.81\%. This enhancement stems from the \textbf{R} pulse, which includes only CN, MCI, and AD samples and employs repeated scans for imaging. The repeated acquisitions yield finer-grained image details that help classify samples ambiguous in the base \textbf{G} pulse. In practice, resampling with other pulse sequence improves tissue sensitivity in regions affected by motion artifacts, which enhances the ability of model to discriminate these challenging classes.

\subsubsection{Analysis of \textbf{GSR} Results}
When the three pulse types are combined, CEPerFed achieves an average accuracy of 88.82\%. Benefiting from the local and global collaborative optimization module, the method also attains a minimal standard deviation of $\pm$1.36, indicating highly stable convergence. Notably, nearly all compared methods exhibit accuracy gains under the
\textbf{GSR} pulse combination, indicating that our multi-pulse MRI design can significantly enhance disease classification performance.

\subsubsection{Performance Comparison across Pulse Combinations}
\begin{figure}[htb]
\centering
\includegraphics[width=\linewidth]{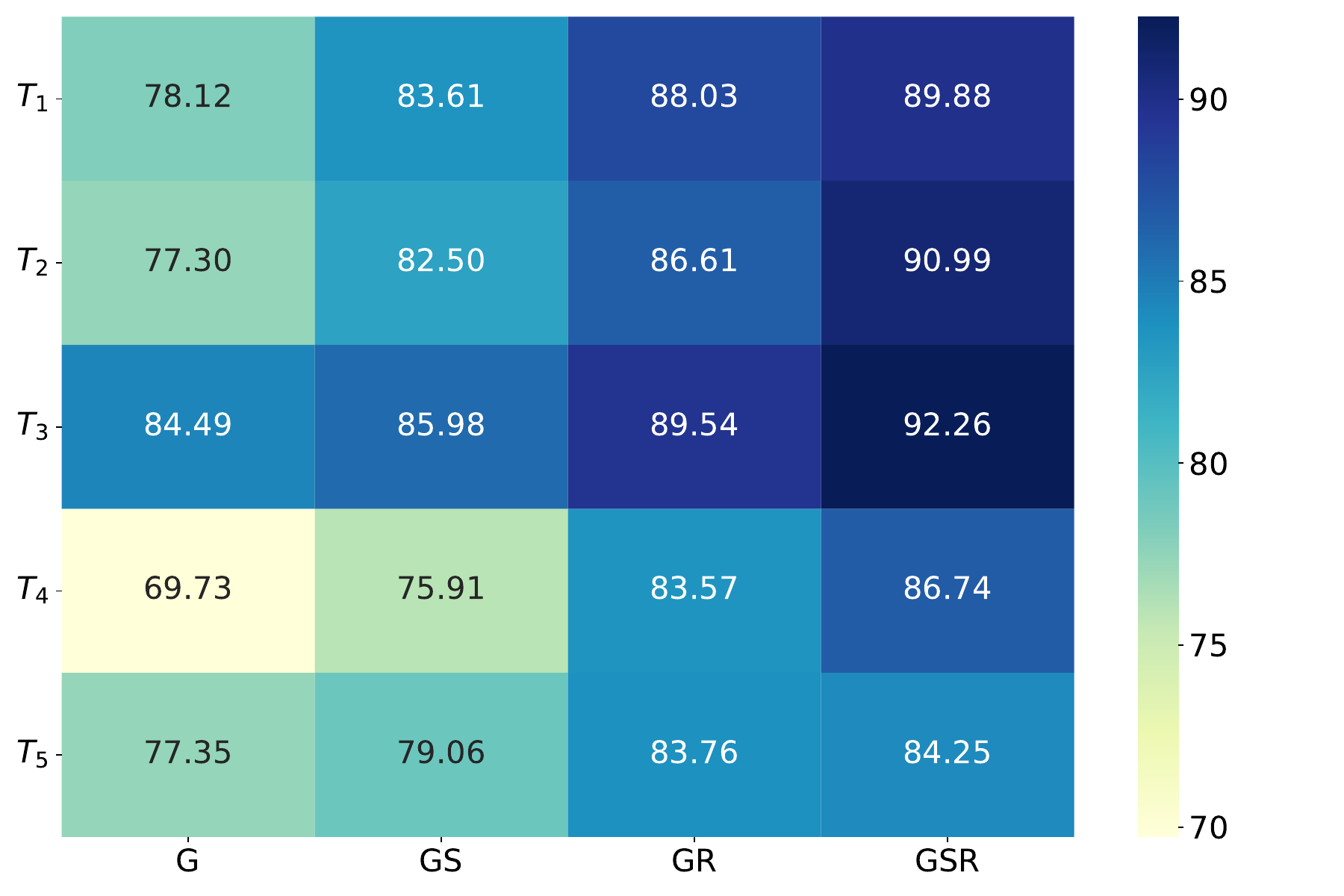}
\caption{Heatmap showing the average classification accuracy across the five diagnostic tasks for each pulse configuration.}
\label{fig:hot_fig}
\end{figure}

To provide a more intuitive comparison of experimental results for pulse MRI combinations across classification tasks, we present the heatmap in Fig.~\ref{fig:hot_fig}, where the four pulses combinations are shown along the horizontal axis and the five classification tasks along the vertical axis. Each cell displays the average classification accuracy (\%) for that pairing, with darker shades indicating higher performance. 

It is evident that the \textbf{GR} configuration consistently outperforms \textbf{GS}, primarily because roughly one-third of the samples in the \textbf{S} pulse imaging belong to the EMCI and LMCI categories, which are relatively easier to distinguish, thus limiting the potential performance gains. In contrast, the \textbf{GSR} configuration, which incorporates all three pulses, consistently achieves the highest average accuracy, which confirms that the \textbf{S} and \textbf{R} pulses provide complementary information that benefits classification. The \textbf{S} pulse provides complementary imaging viewpoints and enhanced tissue contrast, helping to clarify anatomical boundaries that appear blurred in the \textbf{G} pulse. The \textbf{R} pulse uses repeated acquisitions to mitigate motion-induced artifacts in affected samples, thereby reducing their negative impact on classification.

\subsubsection{Analysis of Non-IID Results}
To simulate data heterogeneity across medical centers (non-IID data), we partitioned the dataset using a Dirichlet distribution with concentration parameter $\alpha=0.5$ and randomly assigned the resulting subsets to clients, introducing distribution skew and sample imbalance. Under this non-IID setup we evaluated the best performing \textbf{GSR} pulse combination. Our method achieved a mean accuracy of 70.06\% with a standard deviation of $\pm$2.06, indicating that the method maintains good convergence and stability under strong distributional shift. Notably, the three-class task \textbf{T\textsubscript{4}} accuracy dropped from 86.74\% to 61.86\% (a decrease of 24.88\%), which we attribute to extreme class imbalance caused by the Dirichlet split. For example, individual client has MCI:CN ratios as skewed as 5:839, which severely impairs discrimination for the MCI class.

Compared to other methods, while FedNP performed comparably to our method under IID settings, its mean accuracy degraded markedly by 28.37\% under non-IID conditions, due to its reliance on global feature estimation without inter client collaboration. In contrast, TurboSVM and FedAu exhibited smaller fluctuations, attributable to their already limited performance under the IID setting and focus on simpler class separability. Compared with FedAvg, CEPerFed increases mean accuracy by 8.19\% and achieves a low standard deviation of $\pm$2.06, indicating improved convergence under heterogeneity. Furthermore, it exhibits stronger robustness under non-IID settings than personalized FL methods like FedALA and PGFed. These advantages are attributed to our local and global collaborative optimization strategy, which enhances stability and generalization by harmonizing personalized and consistent updates.



\begin{figure}[htb]
  \centering
  \subfigure[Accuracy curves for tasks \textbf{T\textsubscript{2}} and \textbf{T\textsubscript{3}} with and w/o risk matrix $\alpha$.]
    {\includegraphics[width=0.49\linewidth,scale=1.0]{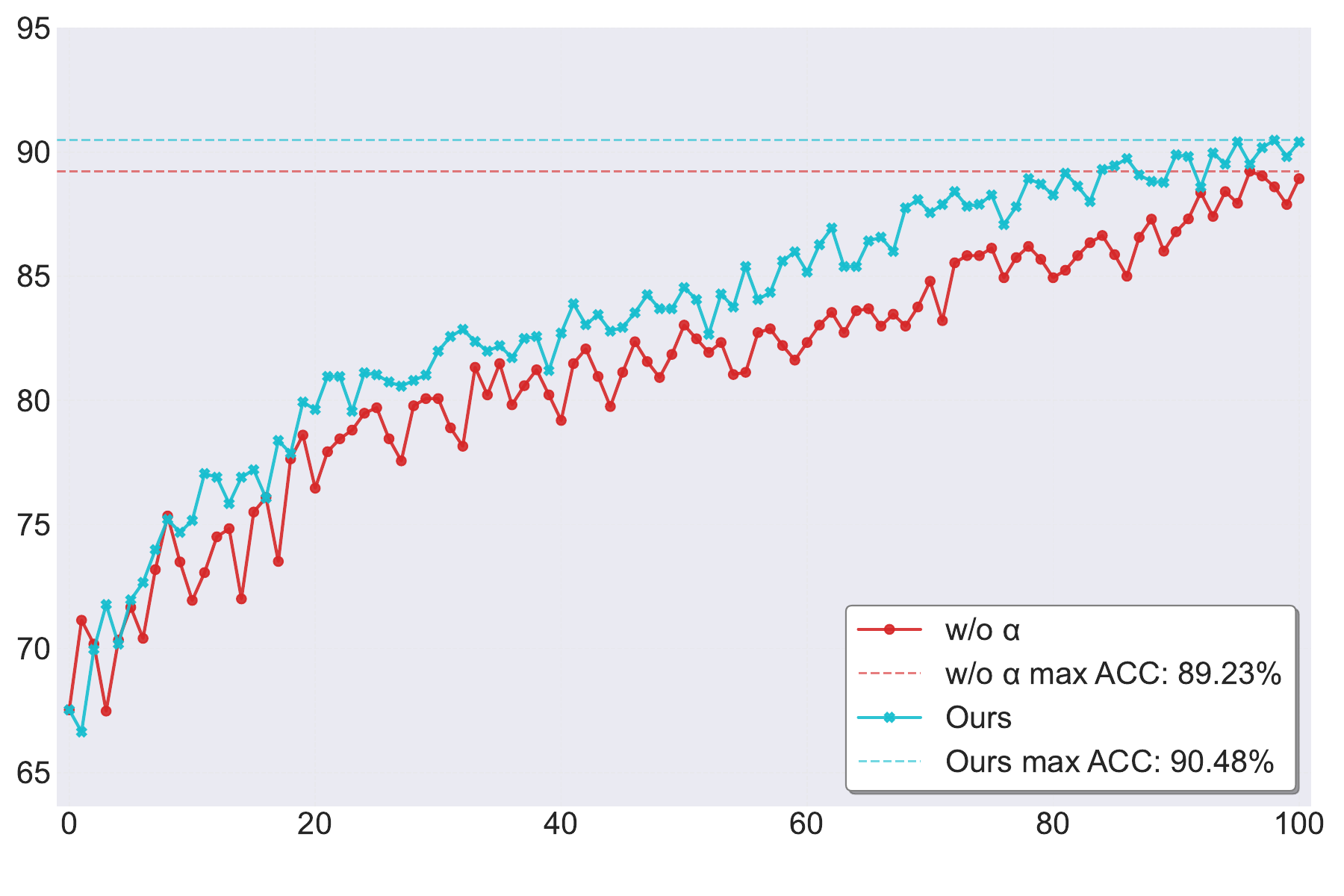}
    \includegraphics[width=0.49\linewidth,scale=1.0]{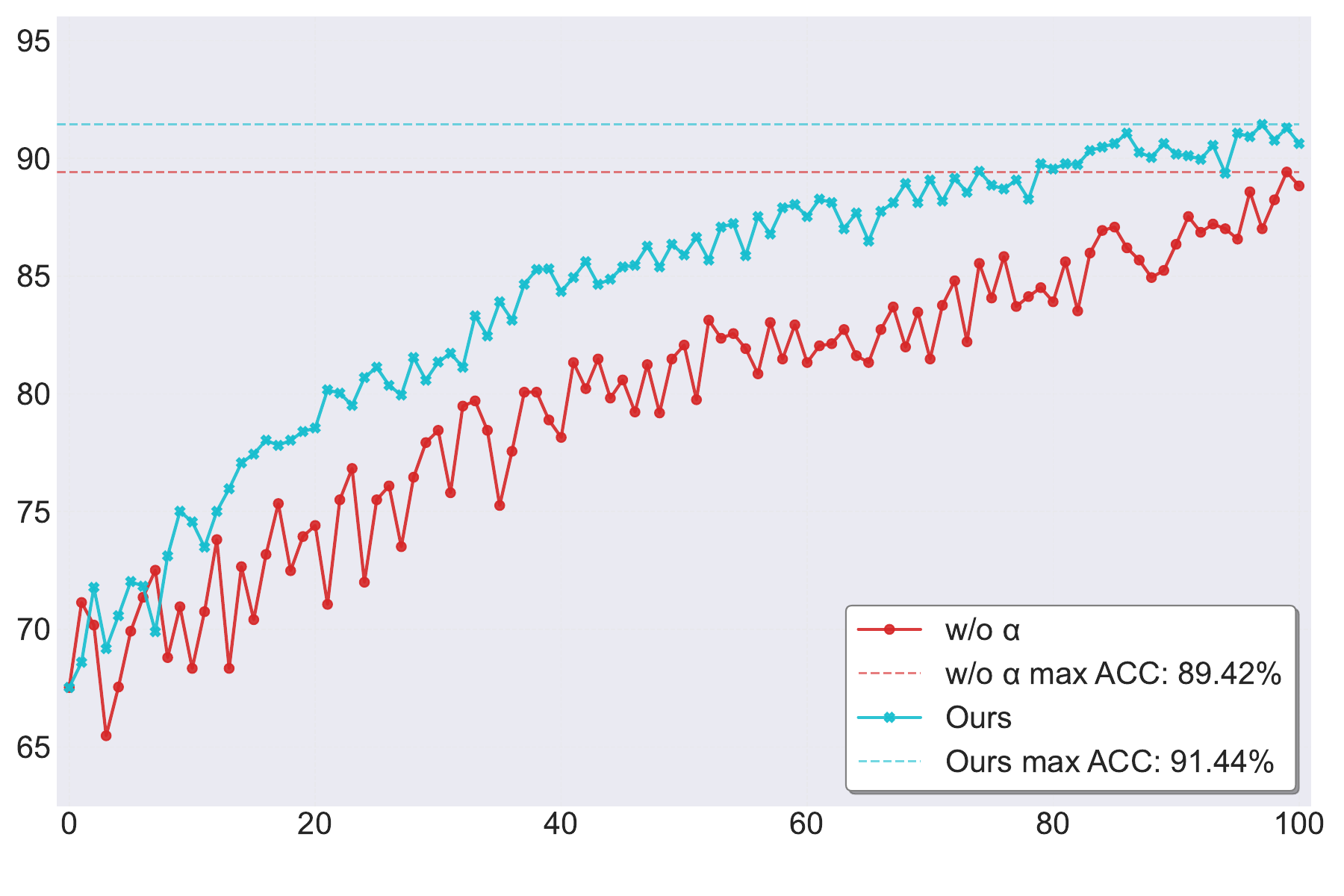}}

    \subfigure[Performance comparison of the five performance metrics for the with and without HSVD strategy on tasks \textbf{T\textsubscript{2}} and \textbf{T\textsubscript{3}}.]
    {\includegraphics[width=0.49\linewidth,scale=1.0]{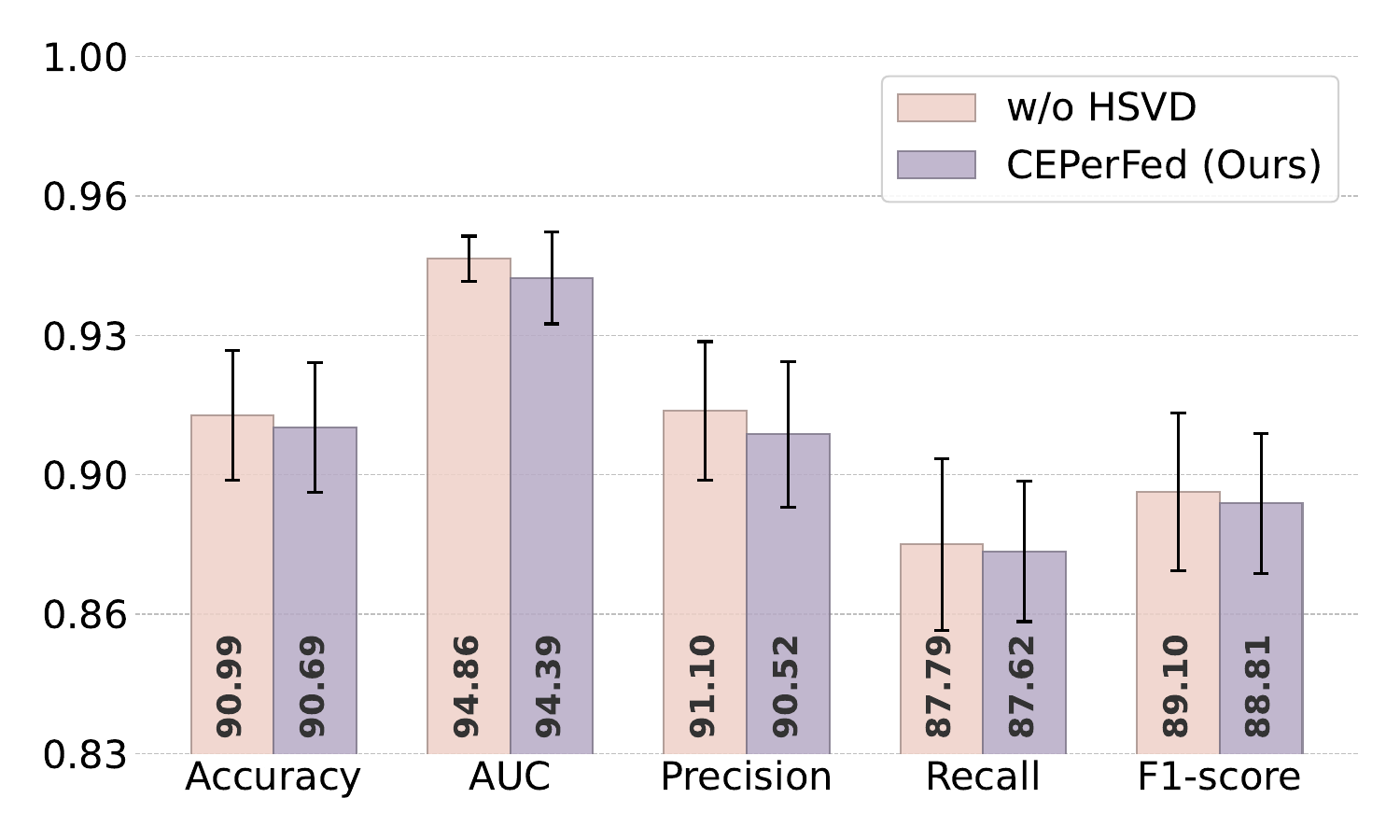}
    \includegraphics[width=0.49\linewidth,scale=1.0]{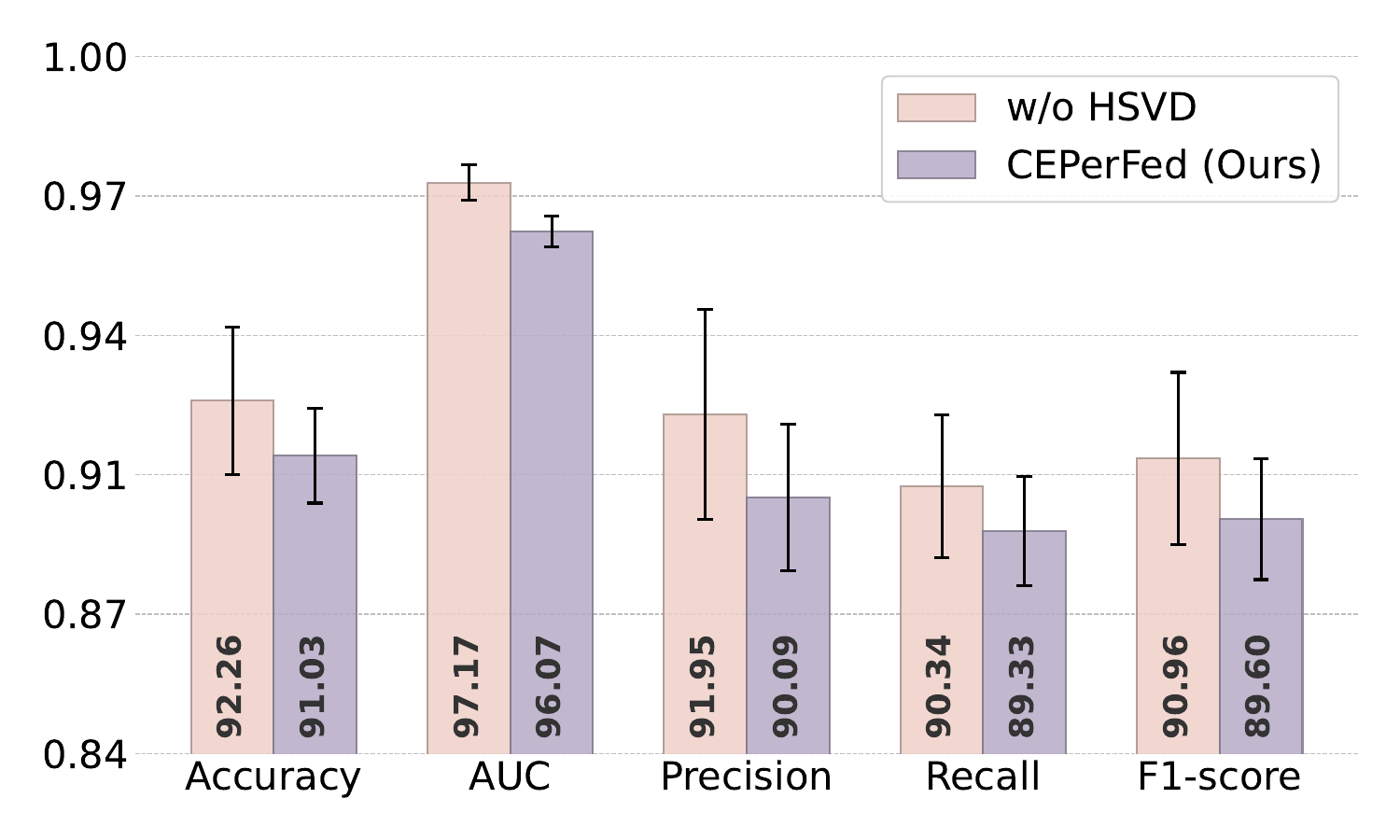}}
  \caption{Ablation experiment results of risk matrix $\alpha$ and HSVD strategy.}
  \label{fig:ablation}
\end{figure}

\subsection{Ablation study}

To evaluate the role of the risk matrix $\alpha$ in our local and global collaborative optimization, we conduct ablation experiments on \textbf{T\textsubscript{2}} (MCI vs. AD) and \textbf{T\textsubscript{3}} (CN vs. AD) tasks. Fig.~\ref{fig:ablation}(a) illustrates the test accuracy trends over the first 100 communication rounds. In the w/o $\alpha$ variant the risk matrix is replaced by uniform weights $1/n$, while all other experimental settings remain unchanged. Over the first 100 communication rounds, our method improves accuracy by 1.25\% on \textbf{T\textsubscript{2}} and by 2.02\% on \textbf{T\textsubscript{3}}. Notably, the accuracy curve for the w/o $\alpha$ variant shows much larger oscillations on \textbf{T\textsubscript{3}}. Introducing the risk matrix reduces these fluctuations during training and produces a clearer accuracy gain. Concretely, the risk matrix assigns larger weights to updates that are beneficial for client. These weights are aggregated into a historical risk gradient, which is then used to correct the raw model gradient.

Additionally, we compared our method with the variant w/o HSVD across five metrics (Fig.~\ref{fig:ablation} (b)). Notably, although redundant parameters are discarded to reduce communication, our method preserves overall performance. This robustness is primarily attributable to the dynamic rank selection strategy. By retaining the top $r$ principal components according to an energy-based criterion while discarding redundant or potentially harmful elements. This strategy not only improves the efficiency of client-side transmission of local updates but also minimizes the loss in model performance.

\begin{figure}[htb]
  \centering
  \subfigure[Performance on \textbf{T\textsubscript{2}}.]
    {\includegraphics[width=0.49\linewidth]{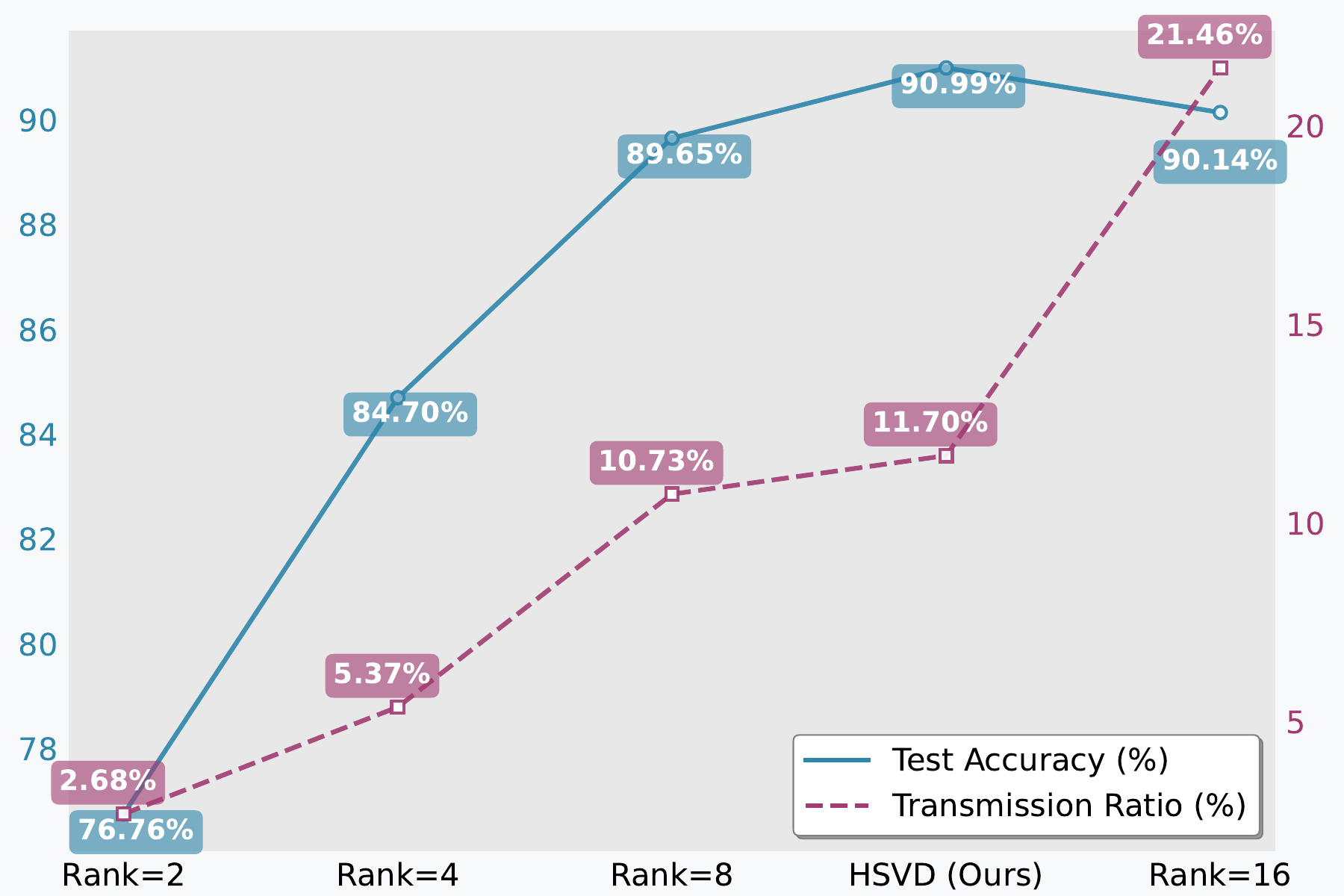}
    \includegraphics[width=0.49\linewidth]{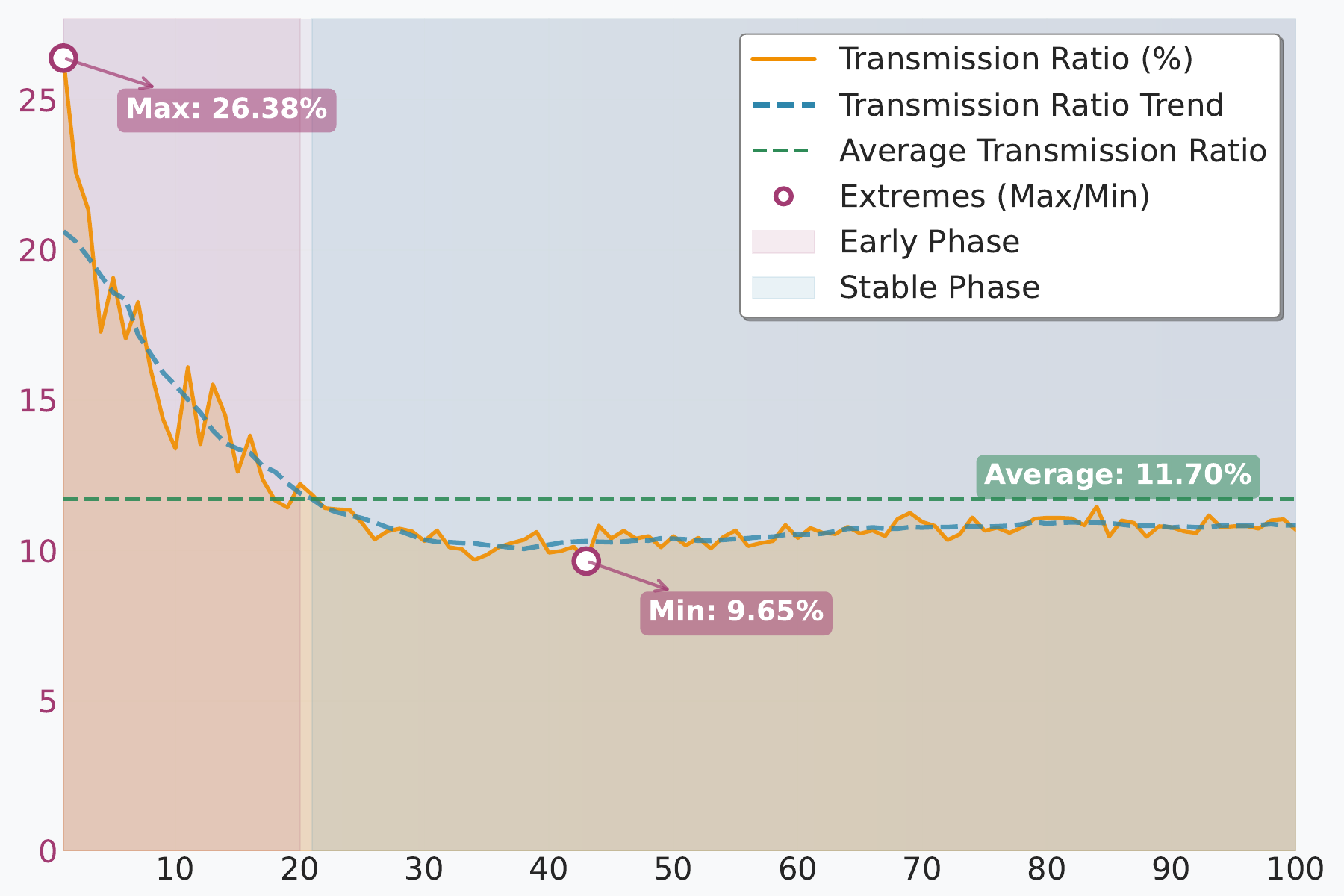}}
    \subfigure[Performance on \textbf{T\textsubscript{3}}.]
    {\includegraphics[width=0.49\linewidth]{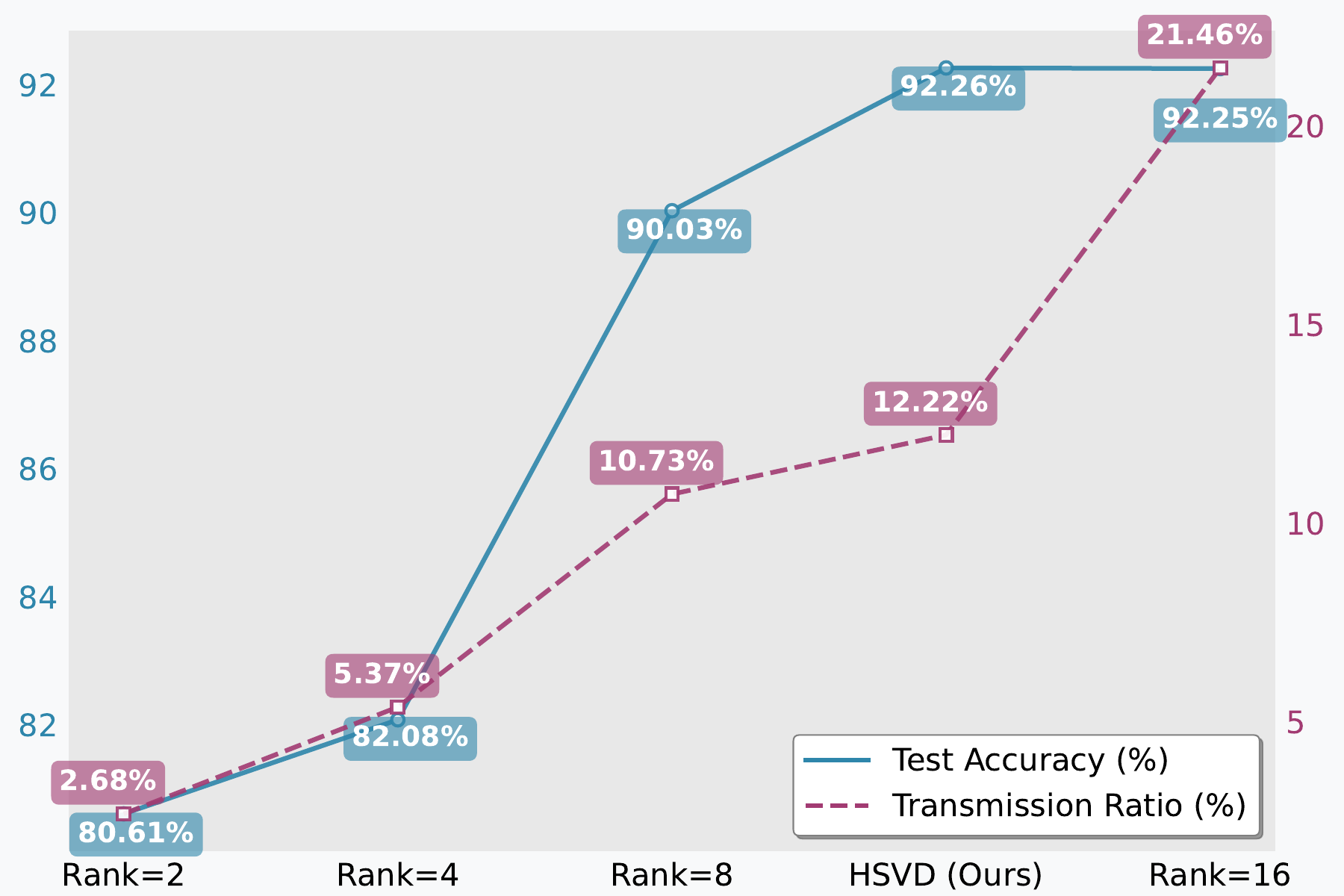}
    \includegraphics[width=0.49\linewidth]{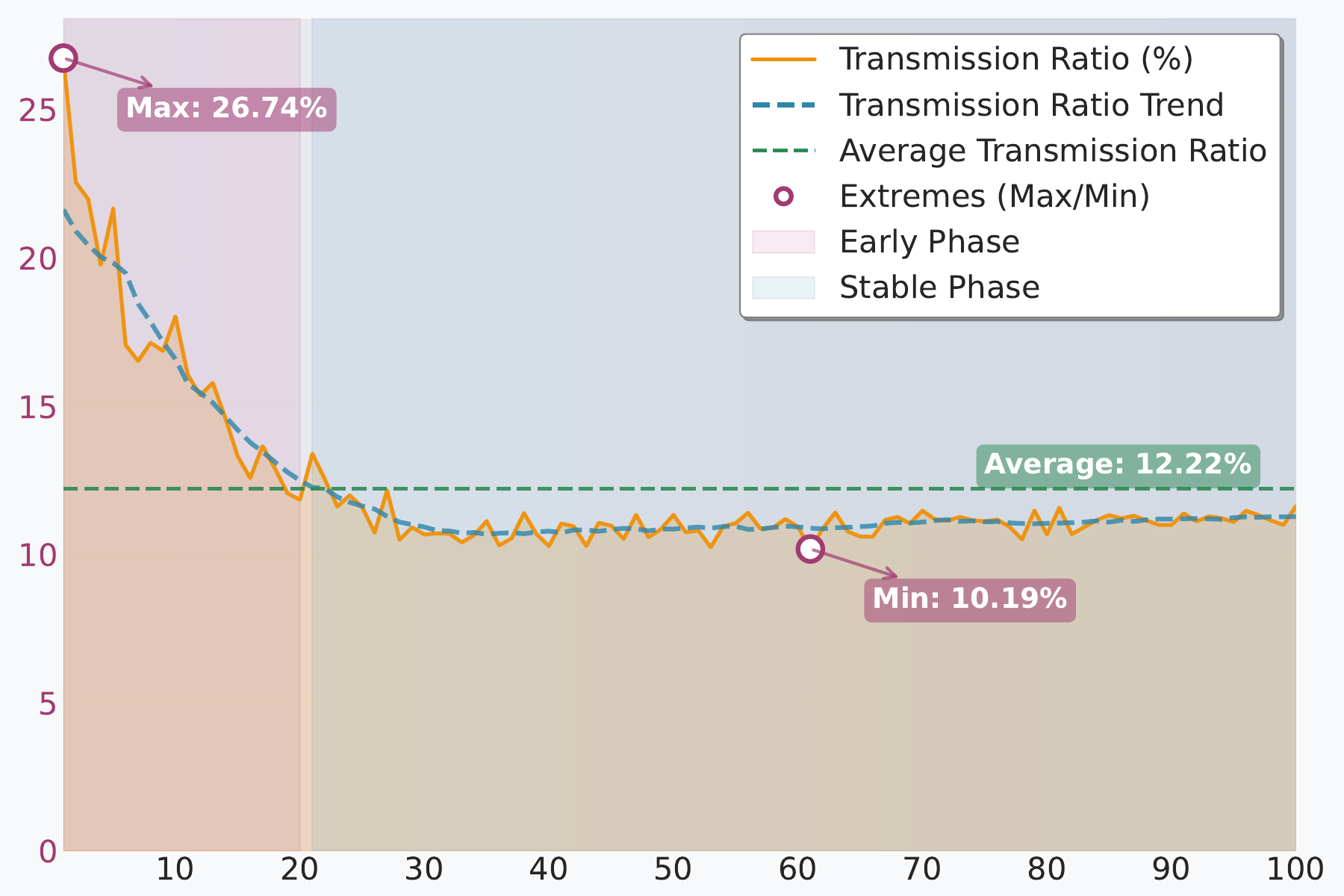}}
  \caption{Comparison of classification accuracy and model parameter transmission ratio between our dynamic rank selection and fixed-rank (left), alongside the trend of the transmission ratio during training (right).}
  \label{fig:all_figures}
\end{figure}

\subsection{Communication Optimization Discussion}
\color{black}
To evaluate the effectiveness of our dynamic rank selection, we compare it with fixed-rank SVD configurations ($r$ = 2, 4, 8, 16) using model transmission ratio and mean classification accuracy. The model transmission ratio is defined as the number of parameters transmitted after decomposition divided by the number of parameters before decomposition. Since the retained rank in our method changes over training, the transmission ratio for the dynamic scheme is the average value over the first 100 communication rounds.

The left column of Fig.~\ref{fig:all_figures} shows that raising the fixed rank improves classification accuracy: low ranks (2 and 4) achieve strong compression but suffer noticeable accuracy loss, while ranks 8 and 16 retain more informative components and yield substantially better performance. Compared with the fixed rank-8 scheme, our dynamic rank transmits only about 1\% more parameters but surpasses the fixed rank-16 configuration in accuracy. The right column of Fig.~\ref{fig:all_figures} plots the transmission ratio across the first 100 rounds, with the blue curve showing a moving average to highlight the trend. The transmission ratio is more variable during the initial 20 rounds while the model is learning and selected ranks are larger. As the model converges, the selected rank and transmission rate tend to be stable.

\section{Conclusion}
In this study, we have proposed a novel communication-efficient PFL method (CEPerFed) for multi-pulse MRI classification. To address the challenges of data heterogeneity and high communication overhead in multi-pulse MRI, we designed a local and global collaborative optimization module and a HSVD strategy, respectively. In the experimental section, we evaluated several pulse combination settings, validating that leveraging multi-pulse MRI configuration substantially improves the diagnostic performance of CEPerFed. Ablation studies and communication optimization analysis further demonstrated the efficacy of each module. To improve client-server communication efficiency from an algorithmic perspective, future work will focus on the development of asynchronous communication algorithm.

\bibliographystyle{IEEEbib}
\bibliography{references}

\end{document}